\documentclass[10pt, a4paper]{article}
\usepackage[table,xcdraw]{xcolor}
\usepackage{lrec-coling2024}  
\usepackage{multirow}
\usepackage{graphicx}
\usepackage{pifont}
\newcommand{\cmark}{\ding{51}}
\usepackage[utf8]{inputenc}
\usepackage{microtype}
\usepackage{inconsolata}
\PassOptionsToPackage{table}{xcolor}
\title{Examining the Limitations of Computational Rumor Detection Models Trained on Static Datasets}

\name{Yida Mu,
     Xingyi Song,
    Kalina Bontcheva,
    Nikolaos Aletras} 

\address{Department of Computer Science, The University of Sheffield\\
         \{y.mu, x.song, k.bontcheva, n.aletras\}@sheffield.ac.uk\\}

\abstract{
A crucial aspect of a rumor detection model is its ability to generalize, particularly its ability to detect emerging, previously unknown rumors. Past research has indicated that content-based (i.e., using solely source post as input) rumor detection models tend to perform less effectively on unseen rumors. At the same time, the potential of context-based models remains largely untapped. The main contribution of this paper is in the in-depth evaluation of the performance gap between content and context-based models specifically on detecting new, unseen rumors. Our empirical findings demonstrate that context-based models are still overly dependent on the information derived from the rumors' source post and tend to overlook the significant role that contextual information can play. We also study the effect of data split strategies on classifier performance. Based on our experimental results, the paper also offers practical suggestions on how to minimize the effects of temporal concept drift in static datasets during the training of rumor detection methods.
 \\ \newline  }

\begin{document}

\maketitleabstract

\section{Introduction}
False rumors are claims or stories that are intended to deceive or mislead the public and can spread faster through social media, causing harm and confusion \citep{lazer2018science,zubiaga2018detection,vosoughi2018spread}. Due to their large volume and high velocity of spread, computational approaches (e.g., supervised rumor detection models) are typically employed to detect and analyze false rumors at an early stage \citep{bian2020rumor,lin2022detect,tian2022duck}.\footnote{Accepted at LREC-COLING 2024.}

Specifically, the task of rumor detection typically distinguishes the detection of check-worthy unverified claims (i.e., rumors) from other kinds of posts in social media (non-rumors) \citep{zubiaga2018detection}. On the other hand, rumor verification\footnote{In this work, for brevity, we refer to both tasks as rumor detection.} is typically the task of classifying a rumor as \textit{True, False, Unverified, or Non-Rumor} \citep{kochkina2023evaluating}. 

Current computational rumor detection systems typically follow a two-step approach: (i) features are extracted from the textual content of the rumor (e.g., source post) along with contextual information,\footnote{In this work, we use the term `contextual information' to refer to different forms of information associated with a rumor on social media, such as comments, images, and user profile attributes. The term 'content-based methods' refers to the use of only source posts as the model input.} and then (ii) models are trained and evaluated on static datasets using random data splits \cite{ma2016detecting,ma2017detect}.

\begin{figure*}[!t]
  \includegraphics[width=15cm]{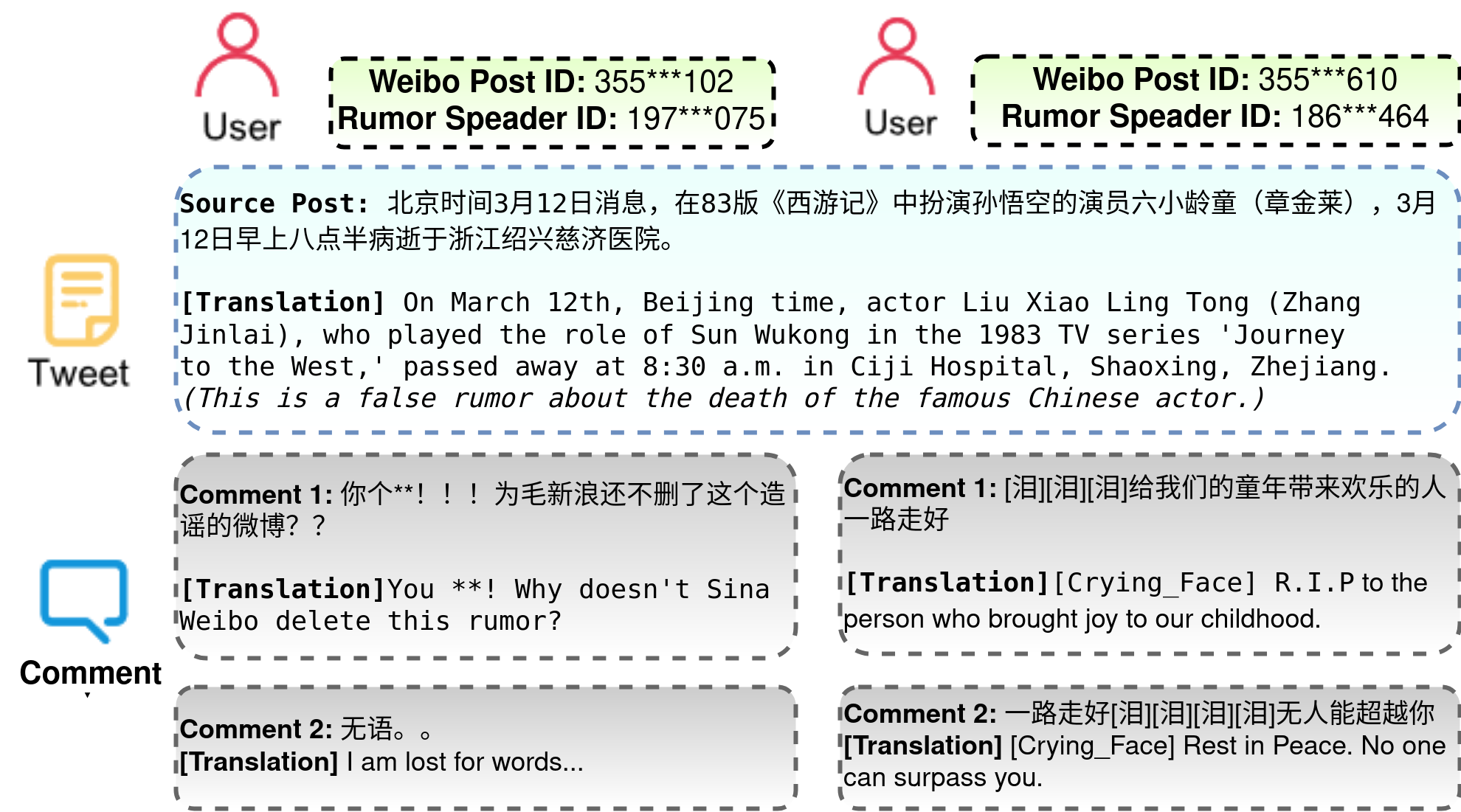}
  \caption{Two rumor spreaders (in the green box) posted an identical rumor and received different stances of comments (in the gray box), i.e., denial (on the left) and support (on the right), respectively. `[Crying\_Face]' denotes the Loudly Crying Face emoji.}
  \label{fig:ex}
\end{figure*}

As demonstrated by \citet{mu2023s,FTT}, the evaluation of rumor detection systems performed on static datasets using random splits might not provide an accurate picture of the generalizability of such models to unseen rumors. Note that the evaluation conducted by \citet{mu2023s,FTT} focuses solely on standard text classifiers (such as logistic regression) using only features derived from source posts. 

However, rumors in social media also come with a rich amount of contextual information, including comments, user profile features and images, which complement the text of the source posts. 
For example, Figure \ref{fig:ex} shows two Weibo users who post the same rumor about the death of a famous Chinese actor. Despite the source posts being identical, the remaining contextual information (e.g., comments and user profile attributes) is completely different. Note that the development of the majority of current rumor detection models relies on context-based features and utilizes random data splits \citep{bian2020rumor,rao-etal-2021-stanker}.

The question that emerges is whether rumor detection models trained with contextual information using random data splits may also exhibit a tendency towards overestimation. Therefore, this paper primarily focuses on a systematic evaluation of the actual generalization capabilities (i.e., detecting rumors that are not previously known) of context-based rumor detection models, which is a hitherto unstudied research question.

The four contributions of this work are:
\begin{itemize}
    \item Empirical proof (\textbf{\S} \ref{datasplits} \& \ref{discussion}) that despite having additional contextual information, rumor detection models still struggle to detect unseen rumors appearing at a future date, with some models performing even worse than random baselines (see Table \ref{tab:results_1}). 
    \item An ablation study (\textbf{\S} \ref{ablation study}) that removes source posts from the inputs, revealed that current rumor detection approaches rely excessively on information from the source post, while neglecting the contextual information. 
    \item A follow-up similarity analysis (\textbf{\S} \ref{section:Similarity}) on content and context-based features, which elucidates the impact of training/test split strategies on model performance.
    \item Finally, we focus on the issue of effectively utilizing static datasets for rumor detection by providing practical recommendations (\textbf{\S} \ref{suggestions}), such as implementing additional cleaning measures for the static dataset and enhancing the current evaluation metrics.
\end{itemize}

\begin{table*}[!t]
\centering
\begin{tabular}{|l|ccccc|}
\hline
\rowcolor[HTML]{c0c0c0} 
\textbf{Statistic} &
  \multicolumn{1}{l|}{\cellcolor[HTML]{c0c0c0}\textbf{Twitter 15}} &
  \multicolumn{1}{l|}{\cellcolor[HTML]{c0c0c0}\textbf{Twitter 16}} &
  \multicolumn{1}{l|}{\cellcolor[HTML]{c0c0c0}\textbf{Weibo 16}} &
  \multicolumn{1}{l|}{\cellcolor[HTML]{c0c0c0}\textbf{Weibo 20}} &
  \multicolumn{1}{l|}{\cellcolor[HTML]{c0c0c0}\textbf{Sun-MM}} \\ \hline
\textit{\# of source posts}         & \multicolumn{1}{c|}{1,490}  & \multicolumn{1}{c|}{818}    & \multicolumn{1}{c|}{4,664}  & \multicolumn{1}{c|}{6,068}  & 2374   \\ \hline
\textit{\# of True rumors}          & \multicolumn{1}{c|}{374}    & \multicolumn{1}{c|}{205}    & \multicolumn{1}{c|}{2,351}  & \multicolumn{1}{c|}{3,034}  & 1,688  \\ \hline
\textit{\# of False rumors}         & \multicolumn{1}{c|}{370}    & \multicolumn{1}{c|}{205}    & \multicolumn{1}{c|}{2,313}  & \multicolumn{1}{c|}{3,034}  & 686    \\ \hline
\textit{\# of Unverified rumors}    & \multicolumn{1}{c|}{374}    & \multicolumn{1}{c|}{203}    & \multicolumn{1}{c|}{-}      & \multicolumn{1}{c|}{-}      & -      \\ \hline
\textit{\# of Non-rumors}           & \multicolumn{1}{c|}{372}    & \multicolumn{1}{c|}{205}    & \multicolumn{1}{c|}{-}      & \multicolumn{1}{c|}{-}      & -      \\ \hline
\textit{Average length of posts}    & \multicolumn{1}{c|}{19}     & \multicolumn{1}{c|}{19}     & \multicolumn{1}{c|}{105}    & \multicolumn{1}{c|}{88}     & -      \\ \hline
\textit{Average \# of comments}     & \multicolumn{1}{c|}{22}     & \multicolumn{1}{c|}{16}     & \multicolumn{1}{c|}{804}    & \multicolumn{1}{c|}{62}     & -      \\ \hline
\textit{Average length of comments} & \multicolumn{1}{c|}{242}    & \multicolumn{1}{c|}{202}    & \multicolumn{1}{c|}{8,484}  & \multicolumn{1}{c|}{13,592} & -      \\ \hline
\rowcolor[HTML]{c0c0c0} 
\textbf{Contextual Information}                 & \multicolumn{5}{l|}{\cellcolor[HTML]{c0c0c0}}                                                                                  \\ \hline
\textit{Source Posts}               & \multicolumn{1}{c|}{\cmark} & \multicolumn{1}{c|}{\cmark} & \multicolumn{1}{c|}{\cmark} & \multicolumn{1}{c|}{\cmark} & \cmark \\ \hline
\textit{Comments}        & \multicolumn{1}{c|}{G}      & \multicolumn{1}{c|}{G}      & \multicolumn{1}{c|}{G+S}    & \multicolumn{1}{c|}{S}    & -      \\ \hline
\textit{User Profile Attributes}    & \multicolumn{1}{c|}{\cmark} & \multicolumn{1}{c|}{\cmark} & \multicolumn{1}{c|}{\cmark} & \multicolumn{1}{c|}{\cmark} & \cmark \\ \hline
\textit{Images}                     & \multicolumn{1}{c|}{-}      & \multicolumn{1}{c|}{-}      & \multicolumn{1}{c|}{-}      & \multicolumn{1}{c|}{-}      & \cmark \\ \hline
\end{tabular}%
% }
\caption{Dataset statistics. `G' and `S' denote comment propagation network (Graph) and comment sequence (S) respectively. We also present contextual-based features obtained from each dataset.}
\label{tab:datasets}
\end{table*}

\section{Related Work}
\subsection{Computational Rumor Detection Approaches}
The increased consumption of news and information on social platforms has necessitated large-scale automated detection of unreliable content \citep{shu2017fake, shearer2017news}, which led to the development of new rumor detection approaches based on state-of-the-art NLP techniques. 

Early studies typically relied on handcrafted features extracted from source posts and user profile attributes using traditional machine learning models, such as SVM and Random Forest. \citep{qazvinian2011rumor,takahashi2012rumor,yang2012automatic,ma2015detect}. With the emergence of neural-based NLP models \citep{Mikolov2013}, rumors started to be modeled with contextual embeddings such as Glove \citep{pennington2014glove} and ELMo \citep{peters-etal-2018-deep}. In addition, graph-based neural models have been employed to learn relationships from the propagation network of rumors, which includes retweet and comment chains \citep{bian2020rumor, lin2021rumor, yang2021rumor}. Other methods adopted multi-modal approaches to go beyond text and capture information from images \citep{wang2020fake,sun-etal-2021-inconsistency-matters,zhou2022mdmn}.

Recent hybrid models began including contextual information to improve rumor prediction performance \citep{lu2020gcan,rao-etal-2021-stanker,tian2022duck}. The top-performing rumor detection systems (e.g., DUCK \citep{tian2022duck}) rely both on contextual information and user-level attributes, with 98 F1-measure on widely used datasets such as Weibo 16 \citep{ma2017detect} and CoAID \citep{cui2020coaid}.

Most of these rumor detection approaches however have a major weakness, as they are trained using random  data splits which ignore a key temporal dimension of rumors and thus tend to overestimate model performance of future unseen rumors  \citep{huang2018examining,sogaard2021we}.

\subsection{The Effect of Temporal Concept Drift in NLP Downstream Tasks}
Previous work on legal, abusive language, COVID-19, and biomedical classification tasks \citep{huang2019neural,huang2018examining,chalkidis2022improved,mu2023examining,jin2023examining} has investigated the sensitivity of classifiers to temporal concept drift (i.e., the deterioration of their performance due to temporal/topic variation) when evaluated on chronological data splits. However, temporal concept drift mainly affects the rumor text (i.e. new unseen topics), as rumors on the same topic posted by different users have different contextual information. \citet{mu2023s} explore the impact of temporal concept drift on rumor detection using standard text classifiers such as logistic regression and fully fine-tuned BERT.

In contrast, this paper performs an extensive empirical evaluation of the effect of temporal concept drift on neural rumor detection models which combine textual and contextual information.

\begin{table*}[!t]
\centering
\small
% \resizebox{\columnwidth}{!}{%
\begin{tabular}{|l|cccc|ccccc|}
\hline
\rowcolor[HTML]{c0c0c0} 
\multicolumn{1}{|c|}{\cellcolor[HTML]{c0c0c0}} &
  \multicolumn{4}{c|}{\cellcolor[HTML]{c0c0c0}\textbf{Contextual Information}} &
  \multicolumn{5}{c|}{\cellcolor[HTML]{c0c0c0}\textbf{Datasets}} \\ \cline{2-10} 
\rowcolor[HTML]{c0c0c0} 
\multicolumn{1}{|c|}{\multirow{-2}{*}{\cellcolor[HTML]{c0c0c0}\textbf{Models}}} &
  \multicolumn{1}{c|}{\cellcolor[HTML]{c0c0c0}\textbf{Post}} &
  \multicolumn{1}{c|}{\cellcolor[HTML]{c0c0c0}\textbf{Comment}} &
  \multicolumn{1}{c|}{\cellcolor[HTML]{c0c0c0}\textbf{User}} &
  \textbf{Image} &
  \multicolumn{1}{c|}{\cellcolor[HTML]{c0c0c0}\textbf{Twitter 15}} &
  \multicolumn{1}{c|}{\cellcolor[HTML]{c0c0c0}\textbf{Twitter 16}} &
  \multicolumn{1}{c|}{\cellcolor[HTML]{c0c0c0}\textbf{Weibo 16}} &
  \multicolumn{1}{c|}{\cellcolor[HTML]{c0c0c0}\textbf{Weibo 20}} &
  \textbf{Sun-MM} \\ \hline
\textit{SVM-HF} &
  \multicolumn{1}{c|}{\cmark} &
  \multicolumn{1}{c|}{-} &
  \multicolumn{1}{c|}{\cmark} &
  - &
  \multicolumn{1}{c|}{\cmark} &
  \multicolumn{1}{c|}{\cmark} &
  \multicolumn{1}{c|}{\cmark} &
  \multicolumn{1}{c|}{\cmark} &
  \cmark \\ \hline
\textit{BERT} &
  \multicolumn{1}{c|}{\cmark} &
  \multicolumn{1}{c|}{-} &
  \multicolumn{1}{c|}{-} &
  - &
  \multicolumn{1}{c|}{\cmark} &
  \multicolumn{1}{c|}{\cmark} &
  \multicolumn{1}{c|}{\cmark} &
  \multicolumn{1}{c|}{\cmark} &
  \cmark \\ \hline
\textit{H-Trans} &
  \multicolumn{1}{c|}{\cmark} &
  \multicolumn{1}{c|}{\cmark} &
  \multicolumn{1}{c|}{-} &
  - &
  \multicolumn{1}{c|}{-} &
  \multicolumn{1}{c|}{-} &
  \multicolumn{1}{c|}{\cmark} &
  \multicolumn{1}{c|}{\cmark} &
  - \\ \hline
\textit{Bi-GCN} &
  \multicolumn{1}{c|}{\cmark} &
  \multicolumn{1}{c|}{\cmark} &
  \multicolumn{1}{c|}{-} &
  - &
  \multicolumn{1}{c|}{\cmark} &
  \multicolumn{1}{c|}{\cmark} &
  \multicolumn{1}{c|}{\cmark} &
  \multicolumn{1}{c|}{-} &
  - \\ \hline
\textit{Hybrid} &
  \multicolumn{1}{c|}{\cmark} &
  \multicolumn{1}{c|}{-} &
  \multicolumn{1}{c|}{-} &
  \cmark &
  \multicolumn{1}{c|}{-} &
  \multicolumn{1}{c|}{-} &
  \multicolumn{1}{c|}{-} &
  \multicolumn{1}{c|}{-} &
  \cmark \\ \hline
\end{tabular}%
% }
\caption{Model details.}
\label{tab:models}
\end{table*}

\section{Experimental Setup}
\subsection{Data}
For comprehensiveness and reliability, our experiments are carried out on five datasets (see Table ~\ref{tab:datasets} for details), which have been widely used in prior rumor detection research \citep{bian2020rumor,rao-etal-2021-stanker,sun-etal-2021-inconsistency-matters,tian2022duck,lin2022detect}: 

\begin{itemize}
    \item \textbf{Twitter 15 \& Twitter 16} \citep{ma2017detect} are two English datasets that include tweets categorized into one of four categories: \textit{True Rumor (T), False Rumor (F), Non-rumor (NR) and Unverified Rumor (U)}. 
    
    \item \textbf{Weibo 16} \citep{ma2017detect} consists of 4,664 Weibo posts in Chinese. It comprises 2,313 \textit{false rumors} debunked by the official Weibo Fact-checking Platform and 2,351 \textit{non-rumors} sourced from mainstream news sources.
    
    \item \textbf{Weibo 20} \citep{rao-etal-2021-stanker} is a Chinese rumor detection dataset similar to Weibo 16. It provides 3,034 \textit{non-rumors} and 3,034 \textit{false rumors} from the same Weibo fact-checking platform as Weibo 16. 
    
    \item \textbf{Sun-MM} \citep{sun-etal-2021-inconsistency-matters} comprises 2,374 annotated tweets (i.e., \textit{rumor or non-rumor}) that cover both textual (i.e., source post) and visual (i.e., image) information. It is typically used for multi-modal rumor detection. 
\end{itemize}

It should be noted that most prior rumor detection models are typically evaluated on two or three datasets only, typically from a specific language. 

\subsection{Models}
Following \citep{kochkina2023evaluating}, we evaluate a number of top-performing rumor detection models.\footnote{Here, we only consider reproducible models with publicly available code and full implementation details. Note that these models have been extensively employed as baselines in prior research \citep{rao-etal-2021-stanker,tian2022duck}}. Each dataset is used to train at least three models, based on the information it provides (see Table \ref{tab:models} for details). 

\textbf{Weak Baseline}
For reference, we provide a weak baseline by randomly generating predictions compared to the ground truth labels of the test set. 

\textbf{SVM-HF (Source Post + User Profile)}
Similar to \citep{yang2012automatic,ma2015detect}, we use a linear SVM model using source posts represented with TF-IDF and various handcrafted features extracted from user profile attributes e.g., number of followers, account status (i.e., whether a verified account or not), number of historical posts, etc.

\textbf{BERT (Source Post)}
In line with previous work \citep{rao-etal-2021-stanker,tian2022duck}, we use solely source posts as input to fine-tune the Bert-base model\footnote{We use bert-base-uncased and bert-base-chinese models from Hugging Face \citep{wolf2020transformers} for English and Chinese datasets respectively.} \citep{devlin2019bert} by adding a linear layer on top of the 12-layer transformer architecture with a softmax activation. We consider the special token `[CLS]' as the post-level representation.

\textbf{Bi-GCN (Comment Network)}
To model the network of comment propagation, we use Bi-Directional Graph Convolutional Networks (Bi-GCN) \citep{bian2020rumor}. Bi-GCN employs two separate GCNs with (i) a top-down directed graph representing rumor spread to learn the patterns of rumor propagation; and (ii) another GCN with an opposite directed graph of rumor diffusion.

\textbf{Hierarchical Transformers (Source Post + Comment Sequence)}
Similar to prior work \citep{rao-etal-2021-stanker,tian2022duck}, we use a hierarchical transformer-based network to encode separately the source post and its sequence of comments.\footnote{Given that the total number of tokens of the source post and all comments exceeds the maximum input length (i.e., 512 tokens) of most Bert-style models.} We then add a self-attention and a linear projection layer with softmax activation to combine the hidden representation of posts and comments.

\textbf{Hybrid Vision-and-Language Representation (Source Post + Image)}
We use visual transformer\footnote{\url{https://huggingface.co/google/vit-base-patch16-224}} (ViT) \citep{dosovitskiy2020image} and BERT \citep{devlin2019bert} to represent images and source posts of rumors for the Sun-MM dataset. We then combine the two hidden representations by adding a fully connected layer with softmax activation for rumor classification.

\subsection{Data Pre-processing}
We begin by processing all the source posts and comments, replacing @mentions and links with special tokens such as `@USR' and `URL' respectively. For the English datasets, we also convert all tweets to lowercase before feeding them to the bert-base-uncased model. 
% Furthermore, all user profile attributes are normalized prior to their usage.

\subsection{Evaluation Metrics}
We run each model three times with different random seeds. In accordance with the original settings \citep{ma2016detecting,ma2017detect,rao-etal-2021-stanker}, we report the average macro precision, recall, F1-score, and accuracy for all binary datasets, i.e., Weibo 16, Weibo 20, and Sun-MM. Since the Twitter datasets (Twitter 15 \& 16) have multi-class labels, we report the average accuracy and F1-score for each class.

\subsection{Hyper-parameters}
For linear SVM, we use word-level and character-level tokenizers for English and Chinese datasets respectively. 
We set learn rate as 2e-5 and batch size as 16 for the Bert-base model. For all transformer-based models, we set the max input length as 512 covering all posts. The implementation details of Bi-GCN are available from the open-source repositories.\footnote{\url{https://github.com/TianBian95/BiGCN}}
All experiments are performed using a single Nvidia RTX Titan GPU with 24GB memory.

% Implementation details (e.g., hyper-parameters) are provided in Appendix \ref{sec:appendix}.

% In accordance with the original settings \citep{ma2016detecting,ma2017detect,rao-etal-2021-stanker}, we report the average macro precision, recall, F1-score, and accuracy for all binary datasets, i.e., Weibo 16, Weibo 20, and Sun-MM. Since the Twitter datasets (Twitter 15 \& 16) have multi-class labels, we report the average accuracy and F1-score for each class. 

\begin{table*}[!t]
\centering
% \small
\resizebox{\textwidth}{!}{%
\begin{tabular}{|ll|ccccc|ccccc|}
\hline
\rowcolor[HTML]{c0c0c0} 
\multicolumn{2}{|c|}{\cellcolor[HTML]{c0c0c0}} &
  \multicolumn{5}{c|}{\cellcolor[HTML]{c0c0c0}\textbf{Twitter 15}} &
  \multicolumn{5}{c|}{\cellcolor[HTML]{c0c0c0}\textbf{Twitter 16}} \\ \cline{3-12} 
\rowcolor[HTML]{c0c0c0} 
\multicolumn{2}{|c|}{\cellcolor[HTML]{c0c0c0}} &
  \multicolumn{1}{l|}{\cellcolor[HTML]{c0c0c0}} &
  \multicolumn{1}{c|}{\cellcolor[HTML]{c0c0c0}\textit{\textbf{NR}}} &
  \multicolumn{1}{c|}{\cellcolor[HTML]{c0c0c0}\textit{\textbf{F}}} &
  \multicolumn{1}{c|}{\cellcolor[HTML]{c0c0c0}\textit{\textbf{T}}} &
  \textit{\textbf{U}} &
  \multicolumn{1}{l|}{\cellcolor[HTML]{c0c0c0}} &
  \multicolumn{1}{c|}{\cellcolor[HTML]{c0c0c0}\textit{\textbf{NR}}} &
  \multicolumn{1}{c|}{\cellcolor[HTML]{c0c0c0}\textit{\textbf{F}}} &
  \multicolumn{1}{c|}{\cellcolor[HTML]{c0c0c0}\textit{\textbf{T}}} &
  \textit{\textbf{U}} \\ \cline{4-7} \cline{9-12} 
\rowcolor[HTML]{c0c0c0} 
\multicolumn{2}{|c|}{\multirow{-3}{*}{\cellcolor[HTML]{c0c0c0}\textbf{Models \& Splits}}} &
  \multicolumn{1}{l|}{\multirow{-2}{*}{\cellcolor[HTML]{c0c0c0}\textit{\textbf{Acc.}}}} &
  \multicolumn{1}{c|}{\cellcolor[HTML]{c0c0c0}\textit{\textbf{F1}}} &
  \multicolumn{1}{c|}{\cellcolor[HTML]{c0c0c0}\textit{\textbf{F1}}} &
  \multicolumn{1}{c|}{\cellcolor[HTML]{c0c0c0}\textit{\textbf{F1}}} &
  \textit{\textbf{F1}} &
  \multicolumn{1}{l|}{\multirow{-2}{*}{\cellcolor[HTML]{c0c0c0}\textit{\textbf{Acc.}}}} &
  \multicolumn{1}{c|}{\cellcolor[HTML]{c0c0c0}\textbf{F1}} &
  \multicolumn{1}{c|}{\cellcolor[HTML]{c0c0c0}\textbf{F1}} &
  \multicolumn{1}{c|}{\cellcolor[HTML]{c0c0c0}\textbf{F1}} &
  \textbf{F1} \\ \hline
  \multicolumn{2}{|l|}{\textbf{Weak Baseline}} &
  \multicolumn{1}{l|}{0.240} &
  \multicolumn{1}{l|}{0.224} &
  \multicolumn{1}{l|}{0.246} &
  \multicolumn{1}{l|}{0.238} &
  \multicolumn{1}{l|}{0.254} &
  \multicolumn{1}{l|}{0.248} &
  \multicolumn{1}{l|}{0.174} &
  \multicolumn{1}{l|}{0.250} &
  \multicolumn{1}{l|}{0.300} &
  \multicolumn{1}{l|}{0.264} \\ \hline
\multicolumn{1}{|l|}{} &
  \textit{Random} &
  \multicolumn{1}{c|}{\cellcolor[HTML]{EFEFEF}\textbf{0.739}} &
  \multicolumn{1}{c|}{\cellcolor[HTML]{EFEFEF}\textbf{0.727}} &
  \multicolumn{1}{c|}{\cellcolor[HTML]{EFEFEF}\textbf{0.701}} &
  \multicolumn{1}{c|}{\cellcolor[HTML]{EFEFEF}\textbf{0.803}} &
  \cellcolor[HTML]{EFEFEF}\textbf{0.728} &
  \multicolumn{1}{c|}{\cellcolor[HTML]{EFEFEF}\textbf{0.709}} &
  \multicolumn{1}{c|}{\cellcolor[HTML]{EFEFEF}\textbf{0.697}} &
  \multicolumn{1}{c|}{\cellcolor[HTML]{EFEFEF}\textbf{0.602}} &
  \multicolumn{1}{c|}{\cellcolor[HTML]{EFEFEF}\textbf{0.858}} &
  \cellcolor[HTML]{EFEFEF}\textbf{0.663} \\ \cline{2-12} 
\multicolumn{1}{|l|}{} &
  \textit{Forward} &
  \multicolumn{1}{c|}{0.413} &
  \multicolumn{1}{c|}{0.589} &
  \multicolumn{1}{c|}{0.366} &
  \multicolumn{1}{c|}{0.092} &
  0.304 &
  \multicolumn{1}{c|}{0.373} &
  \multicolumn{1}{c|}{0.523} &
  \multicolumn{1}{c|}{0.226} &
  \multicolumn{1}{c|}{0.297} &
  0.214 \\ \cline{2-12} 
\multicolumn{1}{|l|}{\multirow{-3}{*}{\textbf{SVM-HF}}} &
  \textit{Reverse} &
  \multicolumn{1}{c|}{0.353} &
  \multicolumn{1}{c|}{0.590} &
  \multicolumn{1}{c|}{0.462} &
  \multicolumn{1}{c|}{0.063} &
  0.062 &
  \multicolumn{1}{c|}{0.380} &
  \multicolumn{1}{c|}{0.520} &
  \multicolumn{1}{c|}{0.103} &
  \multicolumn{1}{c|}{0.411} &
  0.368 \\ \hline
\multicolumn{1}{|l|}{} &
  \textit{Random} &
  \multicolumn{1}{c|}{\cellcolor[HTML]{EFEFEF}\textbf{0.615}} &
  \multicolumn{1}{c|}{\cellcolor[HTML]{EFEFEF}\textbf{0.561}} &
  \multicolumn{1}{c|}{\cellcolor[HTML]{EFEFEF}\textbf{0.593}} &
  \multicolumn{1}{c|}{\cellcolor[HTML]{EFEFEF}\textbf{0.692}} &
  \cellcolor[HTML]{EFEFEF}\textbf{0.599} &
  \multicolumn{1}{c|}{\cellcolor[HTML]{EFEFEF}\textbf{0.598}} &
  \multicolumn{1}{c|}{0.381} &
  \multicolumn{1}{c|}{\cellcolor[HTML]{EFEFEF}\textbf{0.615}} &
  \multicolumn{1}{c|}{\cellcolor[HTML]{EFEFEF}\textbf{0.698}} &
  \cellcolor[HTML]{EFEFEF}\textbf{0.625} \\ \cline{2-12} 
\multicolumn{1}{|l|}{} &
  \textit{Forward} &
  \multicolumn{1}{c|}{0.366} &
  \multicolumn{1}{c|}{0.382} &
  \multicolumn{1}{c|}{0.226} &
  \multicolumn{1}{c|}{0.457} &
  0.328 &
  \multicolumn{1}{c|}{0.380} &
  \multicolumn{1}{c|}{\textbf{0.446}} &
  \multicolumn{1}{c|}{0.306} &
  \multicolumn{1}{c|}{0.110} &
  0.489 \\ \cline{2-12} 
\multicolumn{1}{|l|}{\multirow{-3}{*}{\textbf{BERT}}} &
  \textit{Reverse} &
  \multicolumn{1}{c|}{0.367} &
  \multicolumn{1}{c|}{0.430} &
  \multicolumn{1}{c|}{0.256} &
  \multicolumn{1}{c|}{0.455} &
  0.292 &
  \multicolumn{1}{c|}{0.428} &
  \multicolumn{1}{c|}{0.371} &
  \multicolumn{1}{c|}{0.210} &
  \multicolumn{1}{c|}{0.662} &
  0.483 \\ \hline
\multicolumn{1}{|l|}{} &
  \textit{Random} &
  \multicolumn{1}{c|}{\cellcolor[HTML]{EFEFEF}\textbf{0.838}} &
  \multicolumn{1}{c|}{\cellcolor[HTML]{EFEFEF}\textbf{0.785}} &
  \multicolumn{1}{c|}{\cellcolor[HTML]{EFEFEF}\textbf{0.841}} &
  \multicolumn{1}{c|}{\cellcolor[HTML]{EFEFEF}\textbf{0.886}} &
  \cellcolor[HTML]{EFEFEF}\textbf{0.785} &
  \multicolumn{1}{c|}{\cellcolor[HTML]{EFEFEF}\textbf{0.854}} &
  \multicolumn{1}{c|}{\cellcolor[HTML]{EFEFEF}\textbf{0.745}} &
  \multicolumn{1}{c|}{\cellcolor[HTML]{EFEFEF}\textbf{0.861}} &
  \multicolumn{1}{c|}{\cellcolor[HTML]{EFEFEF}\textbf{0.939}} &
  \cellcolor[HTML]{EFEFEF}\textbf{0.847} \\ \cline{2-12} 
\multicolumn{1}{|l|}{} &
  \textit{Forward} &
  \multicolumn{1}{c|}{0.415} &
  \multicolumn{1}{c|}{0.509} &
  \multicolumn{1}{c|}{0.386} &
  \multicolumn{1}{c|}{0.311} &
  0.319 &
  \multicolumn{1}{c|}{0.489} &
  \multicolumn{1}{c|}{0.551} &
  \multicolumn{1}{c|}{0.381} &
  \multicolumn{1}{c|}{0.401} &
  0.511 \\ \cline{2-12} 
\multicolumn{1}{|l|}{\multirow{-3}{*}{\textbf{Bi-GCN}}} &
  \textit{Reverse} &
  \multicolumn{1}{c|}{0.498} &
  \multicolumn{1}{c|}{0.584} &
  \multicolumn{1}{c|}{0.339} &
  \multicolumn{1}{c|}{0.786} &
  0.118 &
  \multicolumn{1}{c|}{0.517} &
  \multicolumn{1}{c|}{0.502} &
  \multicolumn{1}{c|}{0.413} &
  \multicolumn{1}{c|}{0.667} &
  0.419 \\ \hline
\end{tabular}
}
\caption{Experimental results of Twitter 15 \& 16 datasets across three different data split strategies. Cells in \textbf{bold} indicate the best results from all models. Cells in gray indicate that the model trained using random splits achieves significantly better performance than using both forward and backward chronological splits. ($p$ < 0.05, $t$-test).}
\label{tab:results_1}
\end{table*}
% #c0c0c0
\begin{table*}[!t]
\resizebox{\textwidth}{!}{%
\begin{tabular}{|ll|cccc|cccc|cccc|}
\hline
\rowcolor[HTML]{c0c0c0} 
\multicolumn{1}{|l|}{\cellcolor[HTML]{c0c0c0}} &
  \cellcolor[HTML]{c0c0c0} &
  \multicolumn{4}{c|}{\cellcolor[HTML]{c0c0c0}\textit{\textbf{Weibo 16}}} &
  \multicolumn{4}{c|}{\cellcolor[HTML]{c0c0c0}\textit{\textbf{Weibo 20}}} &
  \multicolumn{4}{c|}{\cellcolor[HTML]{c0c0c0}\textbf{Sun-MM}} \\ \cline{3-14} 
\rowcolor[HTML]{c0c0c0} 
\multicolumn{1}{|l|}{\multirow{-2}{*}{\cellcolor[HTML]{c0c0c0}\textbf{Models}}} &
  \multirow{-2}{*}{\cellcolor[HTML]{c0c0c0}\textbf{Splits}} &
  \multicolumn{1}{l|}{\cellcolor[HTML]{c0c0c0}\textit{\textbf{Acc.}}} &
  \multicolumn{1}{c|}{\cellcolor[HTML]{c0c0c0}\textit{\textbf{P}}} &
  \multicolumn{1}{c|}{\cellcolor[HTML]{c0c0c0}\textit{\textbf{R}}} &
  \textit{\textbf{F1}} &
  \multicolumn{1}{l|}{\cellcolor[HTML]{c0c0c0}\textit{\textbf{Acc.}}} &
  \multicolumn{1}{c|}{\cellcolor[HTML]{c0c0c0}\textit{\textbf{P}}} &
  \multicolumn{1}{c|}{\cellcolor[HTML]{c0c0c0}\textit{\textbf{R}}} &
  \textit{\textbf{F1}} &
  \multicolumn{1}{l|}{\cellcolor[HTML]{c0c0c0}\textit{\textbf{Acc.}}} &
  \multicolumn{1}{c|}{\cellcolor[HTML]{c0c0c0}\textit{\textbf{P}}} &
  \multicolumn{1}{c|}{\cellcolor[HTML]{c0c0c0}\textit{\textbf{R}}} &
  \textit{\textbf{F1}} \\ \hline
\multicolumn{2}{|l|}{\textbf{Weak Baseline}} &
  \multicolumn{1}{l|}{0.493} &
  \multicolumn{1}{l|}{0.493} &
  \multicolumn{1}{l|}{0.492} &
  \multicolumn{1}{l|}{0.493} &
  \multicolumn{1}{l|}{0.501} &
  \multicolumn{1}{l|}{0.501} &
  \multicolumn{1}{l|}{0.501} &
  \multicolumn{1}{l|}{0.501} &
  \multicolumn{1}{l|}{0.514} &
  \multicolumn{1}{l|}{0.512} &
  \multicolumn{1}{l|}{0.514} &
  \multicolumn{1}{l|}{0.512} \\ \hline
\multicolumn{1}{|l|}{} &
  \textit{Random} &
  \multicolumn{1}{c|}{\cellcolor[HTML]{EFEFEF}\textbf{0.906}} &
  \multicolumn{1}{c|}{\cellcolor[HTML]{EFEFEF}\textbf{0.907}} &
  \multicolumn{1}{c|}{\cellcolor[HTML]{EFEFEF}\textbf{0.906}} &
  \cellcolor[HTML]{EFEFEF}\textbf{0.906} &
  \multicolumn{1}{c|}{\cellcolor[HTML]{EFEFEF}\textbf{0.870}} &
  \multicolumn{1}{c|}{\cellcolor[HTML]{EFEFEF}\textbf{0.870}} &
  \multicolumn{1}{c|}{\cellcolor[HTML]{EFEFEF}\textbf{0.868}} &
  \cellcolor[HTML]{EFEFEF}\textbf{0.870} &
  \multicolumn{1}{c|}{\cellcolor[HTML]{EFEFEF}\textbf{0.783}} &
  \multicolumn{1}{c|}{\cellcolor[HTML]{EFEFEF}\textbf{0.742}} &
  \multicolumn{1}{c|}{\cellcolor[HTML]{EFEFEF}\textbf{0.758}} &
  \cellcolor[HTML]{EFEFEF}\textbf{0.749} \\ \cline{2-14} 
\multicolumn{1}{|l|}{} &
  \textit{Forward} &
  \multicolumn{1}{c|}{0.823} &
  \multicolumn{1}{c|}{0.855} &
  \multicolumn{1}{c|}{0.822} &
  0.819 &
  \multicolumn{1}{c|}{0.680} &
  \multicolumn{1}{c|}{0.691} &
  \multicolumn{1}{c|}{0.680} &
  0.676 &
  \multicolumn{1}{c|}{0.689} &
  \multicolumn{1}{c|}{0.635} &
  \multicolumn{1}{c|}{0.630} &
  0.635 \\ \cline{2-14} 
\multicolumn{1}{|l|}{\multirow{-3}{*}{\textbf{SVM-HF}}} &
  \textit{Backward} &
  \multicolumn{1}{c|}{0.752} &
  \multicolumn{1}{c|}{0.757} &
  \multicolumn{1}{c|}{0.752} &
  0.752 &
  \multicolumn{1}{c|}{0.801} &
  \multicolumn{1}{c|}{0.802} &
  \multicolumn{1}{c|}{0.801} &
  0.801 &
  \multicolumn{1}{c|}{0.771} &
  \multicolumn{1}{c|}{0.740} &
  \multicolumn{1}{c|}{0.676} &
  0.692 \\ \hline
\multicolumn{1}{|l|}{} &
  \textit{Random} &
  \multicolumn{1}{c|}{\cellcolor[HTML]{EFEFEF}\textbf{0.918}} &
  \multicolumn{1}{c|}{\cellcolor[HTML]{EFEFEF}\textbf{0.918}} &
  \multicolumn{1}{c|}{\cellcolor[HTML]{EFEFEF}\textbf{0.917}} &
  \cellcolor[HTML]{EFEFEF}\textbf{0.918} &
  \multicolumn{1}{c|}{\cellcolor[HTML]{EFEFEF}\textbf{0.920}} &
  \multicolumn{1}{c|}{\cellcolor[HTML]{EFEFEF}\textbf{0.921}} &
  \multicolumn{1}{c|}{\cellcolor[HTML]{EFEFEF}\textbf{0.920}} &
  \cellcolor[HTML]{EFEFEF}\textbf{0.920} &
  \multicolumn{1}{c|}{\cellcolor[HTML]{EFEFEF}\textbf{0.839}} &
  \multicolumn{1}{c|}{\cellcolor[HTML]{EFEFEF}\textbf{0.807}} &
  \multicolumn{1}{c|}{\cellcolor[HTML]{EFEFEF}\textbf{0.806}} &
  \cellcolor[HTML]{EFEFEF}\textbf{0.806} \\ \cline{2-14} 
\multicolumn{1}{|l|}{} &
  \textit{Forward} &
  \multicolumn{1}{c|}{0.889} &
  \multicolumn{1}{c|}{0.892} &
  \multicolumn{1}{c|}{0.888} &
  0.888 &
  \multicolumn{1}{c|}{0.738} &
  \multicolumn{1}{c|}{0.756} &
  \multicolumn{1}{c|}{0.738} &
  0.732 &
  \multicolumn{1}{c|}{0.708} &
  \multicolumn{1}{c|}{0.682} &
  \multicolumn{1}{c|}{0.708} &
  0.680 \\ \cline{2-14} 
\multicolumn{1}{|l|}{\multirow{-3}{*}{\textbf{BERT}}} &
  \textit{Backward} &
  \multicolumn{1}{c|}{0.809} &
  \multicolumn{1}{c|}{0.812} &
  \multicolumn{1}{c|}{0.809} &
  0.808 &
  \multicolumn{1}{c|}{0.898} &
  \multicolumn{1}{c|}{0.899} &
  \multicolumn{1}{c|}{0.898} &
  0.898 &
  \multicolumn{1}{c|}{0.807} &
  \multicolumn{1}{c|}{0.783} &
  \multicolumn{1}{c|}{0.735} &
  0.748 \\ \hline
\multicolumn{1}{|l|}{} &
  \textit{Random} &
  \multicolumn{1}{c|}{\cellcolor[HTML]{EFEFEF}\textbf{0.892}} &
  \multicolumn{1}{c|}{\cellcolor[HTML]{EFEFEF}\textbf{0.893}} &
  \multicolumn{1}{c|}{\cellcolor[HTML]{EFEFEF}\textbf{0.885}} &
  \cellcolor[HTML]{EFEFEF}\textbf{0.887} &
  \multicolumn{1}{c|}{\textbf{-}} &
  \multicolumn{1}{c|}{\textbf{-}} &
  \multicolumn{1}{c|}{\textbf{-}} &
  \textbf{-} &
  \multicolumn{1}{c|}{-} &
  \multicolumn{1}{c|}{-} &
  \multicolumn{1}{c|}{-} &
  - \\ \cline{2-14} 
\multicolumn{1}{|l|}{} &
  \textit{Forward} &
  \multicolumn{1}{c|}{0.843} &
  \multicolumn{1}{c|}{0.843} &
  \multicolumn{1}{c|}{0.834} &
  0.835 &
  \multicolumn{1}{c|}{\textbf{-}} &
  \multicolumn{1}{c|}{\textbf{-}} &
  \multicolumn{1}{c|}{\textbf{-}} &
  \textbf{-} &
  \multicolumn{1}{c|}{-} &
  \multicolumn{1}{c|}{-} &
  \multicolumn{1}{c|}{-} &
  - \\ \cline{2-14} 
\multicolumn{1}{|l|}{\multirow{-3}{*}{\textbf{Bi-GCN}}} &
  \textit{Backward} &
  \multicolumn{1}{c|}{0.762} &
  \multicolumn{1}{c|}{0.783} &
  \multicolumn{1}{c|}{0.762} &
  0.747 &
  \multicolumn{1}{c|}{\textbf{-}} &
  \multicolumn{1}{c|}{\textbf{-}} &
  \multicolumn{1}{c|}{\textbf{-}} &
  \textbf{-} &
  \multicolumn{1}{c|}{-} &
  \multicolumn{1}{c|}{-} &
  \multicolumn{1}{c|}{-} &
  - \\ \hline
\multicolumn{1}{|l|}{} &
  \textit{Random} &
  \multicolumn{1}{c|}{\cellcolor[HTML]{EFEFEF}\textbf{0.955}} &
  \multicolumn{1}{c|}{\cellcolor[HTML]{EFEFEF}\textbf{0.956}} &
  \multicolumn{1}{c|}{\cellcolor[HTML]{EFEFEF}\textbf{0.955}} &
  \cellcolor[HTML]{EFEFEF}\textbf{0.955} &
  \multicolumn{1}{c|}{\cellcolor[HTML]{EFEFEF}\textbf{0.959}} &
  \multicolumn{1}{c|}{\cellcolor[HTML]{EFEFEF}\textbf{0.960}} &
  \multicolumn{1}{c|}{\cellcolor[HTML]{EFEFEF}\textbf{0.959}} &
  \cellcolor[HTML]{EFEFEF}\textbf{0.959} &
  \multicolumn{1}{c|}{\cellcolor[HTML]{EFEFEF}\bf 0.853} &
  \multicolumn{1}{c|}{\cellcolor[HTML]{EFEFEF}\bf 0.818} &
  \multicolumn{1}{c|}{\cellcolor[HTML]{EFEFEF}\bf 0.829} &
  \cellcolor[HTML]{EFEFEF} \bf 0.823 \\ \cline{2-14} 
\multicolumn{1}{|l|}{} &
  \textit{Forward} &
  \multicolumn{1}{l|}{0.946} &
  \multicolumn{1}{l|}{0.949} &
  \multicolumn{1}{l|}{0.946} &
  \multicolumn{1}{l|}{0.946} &
  \multicolumn{1}{l|}{0.850} &
  \multicolumn{1}{l|}{0.860} &
  \multicolumn{1}{l|}{0.849} &
  \multicolumn{1}{l|}{0.850} &
  \multicolumn{1}{c|}{0.707} &
  \multicolumn{1}{c|}{0.687} &
  \multicolumn{1}{c|}{0.725} &
  0.685 \\ \cline{2-14} 
\multicolumn{1}{|l|}{\multirow{-3}{*}{\textbf{H-Trans / Hybrid}}} &
  \textit{Backward} &
  \multicolumn{1}{l|}{0.792} &
  \multicolumn{1}{l|}{0.833} &
  \multicolumn{1}{l|}{0.785} &
  \multicolumn{1}{l|}{0.793} &
  \multicolumn{1}{l|}{0.940} &
  \multicolumn{1}{l|}{0.938} &
  \multicolumn{1}{l|}{0.935} &
  \multicolumn{1}{l|}{0.938} &
  \multicolumn{1}{c|}{0.821} &
  \multicolumn{1}{c|}{0.782} &
  \multicolumn{1}{c|}{0.805} &
  0.791 \\ \hline
\end{tabular}
}
\caption{Experimental results of Weibo 16 \& 20 and Sun-MM across three different data split strategies. Cells in \textbf{bold} indicate the best results from all models. Cells in gray indicate that the model trained using random splits achieves significantly better performance than using both forward and backward chronological splits. ($p$ < 0.05, $t$-test).}
\label{tab:results_2}
\end{table*}

\begin{figure}[!t]
  \includegraphics[width=\columnwidth]{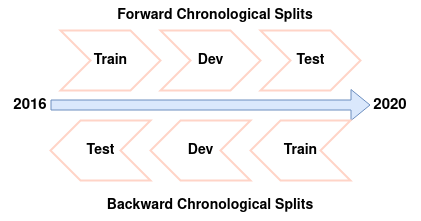}
  \caption{An example of using forward and backward chronological data splits on Weibo 20 dataset (including rumors from 2016 to 2020). There is no overlap among the three subsets.}
  \label{fig:splits}
\end{figure}
\section{Evaluation Strategies}
\subsection{Data Splits}
\label{datasplits}
To examine the effect of data splitting strategies on the models' predictive performance, we compare three strategies: the widely used random data split against two types of chronological data splits (see Figure \ref{fig:splits}). 

\begin{itemize}
    \item \textbf{Forward Chronological Splits}
For each dataset, we initially sort all rumors chronologically, from the oldest to the newest. We then divide them into three subsets: a training set (containing 70\% of the oldest rumors), a development set (10\% of the rumors that were posted after those in the training set but before those in the test set), and a test set (containing the 20\% most recent rumors). This data split strategy allows the model to be trained and fine-tuned on \textit{older rumors} and then be evaluated on the most \textit{recent ones}.
    \item \textbf{Backward Chronological Splits} 
In contrast, here all rumors are sorted starting from the most recent ones to the oldest ones, and then are split in the same way as the forward chronological splits. This allows the model to be trained on the \textit{newest rumors} and evaluated on the \textit{oldest ones}. 
    \item \textbf{Random Splits}
This is the most commonly adopted data split strategy in prior work. All datasets are divided into three subsets using a stratified random split approach\footnote{We use a data split tool from sklearn: \url{https://scikit-learn.org/stable/modules/generated/sklearn.model_selection.train_test_split.html}}. 
\end{itemize}

These two different temporal split strategies enable the evaluation of temporal concept drift effects on model performance.

Some prior rumor detection research has used a leave-a-rumor-out strategy \citep{lukasik2015classifying,lukasik2016hawkes}, where each dataset is divided into $N$ folds, where $N$ denotes the the number of unique rumor events in the given dataset. In this case, rumor detection models are evaluated using N-fold cross validation, i.e., using $N-1$ unique rumors and all associated posts as the training set and the posts about the last remaining rumor as the test set. In this way, it is possible to evaluate model performance on \textit{new unseen rumors}. However, it has not been possible to experiment with this data split protocol as none of the datases used in this paper cluster posts into individual events which give rise to a unique rumor, with associated multiple social media posts about it.

\section{Results and Discussion}
\label{discussion}
\subsection{Model Performance on Random Splits} 
The experimental results for all rumor detection approaches and data split strategies are shown in Tables \ref{tab:results_1} and \ref{tab:results_2}.
We can observe that training on random splits always leads to  significant overestimation (t-test, $p$ < 0.01) of model accuracy as compared to training on both forward and backward chronological splits. 

Taking the best performing Bi-GCN model on Twitter 15 as an example, we observe a decrease in model accuracy of at least 39.4\% when  comparing test results on random splits against the two chronological splits. Furthermore, we find that some models (e.g., SVM-HF and Bi-GCN on Twitter 15) perform even worse than a weak baseline (e.g., the F1-measure results for the false rumor category ($F$) across two chronological splits in comparison with the weak baseline) that uses random predictions. As expected, our empirical findings align with previous studies of temporal impact in other downstream NLP tasks \citep{huang2019neural,chalkidis2022improved,mu2023examining}.

The results indicate that models learn to classify accurately rumor posts in the test set only when they are highly similar to posts in the training data, even though the remaining contextual information (such as user profile attributes, comments, and sometimes images) are different. To further investigate the impact of this semantic overlap, we conduct an ablation study (Section \ref{ablation study}) and a similarity analysis (Section \ref{section:Similarity}).

\subsection{Forward v.s. Backward Chronological Splits}
Our experimental results show that models trained using backward chronological splits achieve higher accuracy on all datasets (except Weibo 16) as compared to those on forward chronological splits. This suggests that the models have the tendency to learn recurrent rumors. This observation is consistent across datasets. For instance, the accuracy of all models on the Twitter 16 dataset is higher when  random splits are used for training as compared to forward splits, but lower when compared to backward splits. This may be attributed to  similarities between the training and test sets. This is investigated further in Section \ref{section:Similarity}.

\subsection{Ablation Study}
\label{ablation study}
In order to evaluate the impact of the source post's text on rumor detection performance, we perform a source post removal ablation study\footnote{Previous ablation studies have focused primarily on removing new features rather than source posts \citep{sun-etal-2021-inconsistency-matters,tian2022duck}}. 
Our hypothesis is that after removing the source posts, there will be no significant difference in the performance of the rumor detection models trained according to the different data split strategies.
We conduct experiments using (i) SVM-HF on all datasets, (ii) the Hier-Transformer model on Weibo 16 and Weibo 20, and (iii) visual transformer (ViT) on Sun-MM dataset.

The results of the ablation study are reported in Table \ref{tab:results_4} and Table \ref{tab:results_5}.
We demonstrate that when the source posts are removed from the input, all models except for ViT model (see Section \ref{section:Similarity} for further analysis) no longer exhibit consistent superiority over forward and backward chronological splits as compared to using random splits. 
As we have shown, two identical rumors can have different contextual information. This indicates that temporalities are not commonly reflected in the majority of contextual information associated with rumors in social media. Notably, even without the source post, the H-Trans model can achieve competitive performance using chronological splits. For instance, it achieves up to 93.8\% and 94.4\% accuracy on Weibo 16 and Weibo 20, respectively, which is comparable to the performance of the Bi-GCN and original H-Trans models (which take the source post as input). 
We hypothesize that rumor debunking information may be present in the comments (for example, see Figure \ref{fig:ex}), which can assist in the decision-making process of the rumor classifier. Next we conduct linguistic analysis to elucidate the distinctions between comments from rumors and non-rumors in Weibo 16 \& 20.

\begin{table*}[!t]
\resizebox{\textwidth}{!}{%
\begin{tabular}{|l|l|ccccl|ccccl|}
\hline
\rowcolor[HTML]{c0c0c0} 
\multicolumn{1}{|c|}{\cellcolor[HTML]{c0c0c0}} &
  \multicolumn{1}{c|}{\cellcolor[HTML]{c0c0c0}} &
  \multicolumn{5}{c|}{\cellcolor[HTML]{c0c0c0}\textit{\textbf{Twitter 15}}} &
  \multicolumn{5}{c|}{\cellcolor[HTML]{c0c0c0}\textit{\textbf{Twitter 16}}} \\ \cline{3-12} 
\rowcolor[HTML]{c0c0c0} 
\multicolumn{1}{|c|}{\multirow{-2}{*}{\cellcolor[HTML]{c0c0c0}\textbf{Models}}} &
  \multicolumn{1}{c|}{\multirow{-2}{*}{\cellcolor[HTML]{c0c0c0}\textbf{Splits}}} &
  \multicolumn{1}{c|}{\cellcolor[HTML]{c0c0c0}\textit{\textbf{Acc.}}} &
  \multicolumn{1}{c|}{\cellcolor[HTML]{c0c0c0}\textit{\textbf{NR}}} &
  \multicolumn{1}{c|}{\cellcolor[HTML]{c0c0c0}\textit{\textbf{F}}} &
  \multicolumn{1}{c|}{\cellcolor[HTML]{c0c0c0}\textit{\textbf{T}}} &
  \multicolumn{1}{c|}{\cellcolor[HTML]{c0c0c0}\textbf{U}} &
  \multicolumn{1}{c|}{\cellcolor[HTML]{c0c0c0}\textit{\textbf{Acc.}}} &
  \multicolumn{1}{c|}{\cellcolor[HTML]{c0c0c0}\textit{\textbf{NR}}} &
  \multicolumn{1}{c|}{\cellcolor[HTML]{c0c0c0}\textit{\textbf{F}}} &
  \multicolumn{1}{c|}{\cellcolor[HTML]{c0c0c0}\textit{\textbf{T}}} &
  \multicolumn{1}{c|}{\cellcolor[HTML]{c0c0c0}\textbf{U}} \\ \hline
 &
  \textit{Random} &
  \multicolumn{1}{c|}{\cellcolor[HTML]{EFEFEF}\textbf{0.383}} &
  \multicolumn{1}{c|}{\cellcolor[HTML]{EFEFEF}0.609} &
  \multicolumn{1}{c|}{\cellcolor[HTML]{EFEFEF}0.050} &
  \multicolumn{1}{c|}{\cellcolor[HTML]{EFEFEF}0.356} &
  \cellcolor[HTML]{EFEFEF}\textbf{0.132} &
  \multicolumn{1}{c|}{\cellcolor[HTML]{EFEFEF}0.343} &
  \multicolumn{1}{c|}{\cellcolor[HTML]{EFEFEF}0.494} &
  \multicolumn{1}{c|}{\cellcolor[HTML]{EFEFEF}0.140} &
  \multicolumn{1}{c|}{\cellcolor[HTML]{EFEFEF}\textbf{0.273}} &
  \cellcolor[HTML]{EFEFEF}\textbf{0.229} \\ \cline{2-12} 
 &
  \textit{Forward} &
  \multicolumn{1}{c|}{0.375} &
  \multicolumn{1}{c|}{\textbf{0.635}} &
  \multicolumn{1}{c|}{0.039} &
  \multicolumn{1}{c|}{\textbf{0.374}} &
  0.086 &
  \multicolumn{1}{c|}{\textbf{0.417}} &
  \multicolumn{1}{c|}{\textbf{0.689}} &
  \multicolumn{1}{c|}{\textbf{0.333}} &
  \multicolumn{1}{c|}{0.046} &
  0.158 \\ \cline{2-12} 
\multirow{-3}{*}{\textbf{\begin{tabular}[c]{@{}l@{}}SVM-HF \\ w/o SP\end{tabular}}} &
  \textit{Reverse} &
  \multicolumn{1}{c|}{0.361} &
  \multicolumn{1}{c|}{0.590} &
  \multicolumn{1}{c|}{\textbf{0.133}} &
  \multicolumn{1}{c|}{0.359} &
  0.050 &
  \multicolumn{1}{c|}{0.328} &
  \multicolumn{1}{c|}{0.499} &
  \multicolumn{1}{c|}{0.178} &
  \multicolumn{1}{c|}{0.170} &
  0.021 \\ \hline
\end{tabular}
}
\caption{Ablation study of Twitter 15 \& 16 datasets across three different data split strategies. Cells in \textbf{bold} indicate the best results from all models.}
\label{tab:results_5}
\end{table*}

\begin{table*}[!t]
\resizebox{\textwidth}{!}{%
\begin{tabular}{|c|l|cccc|cccc|cccc|}
\hline
\rowcolor[HTML]{c0c0c0} 
\multicolumn{1}{|l|}{\cellcolor[HTML]{c0c0c0}} &
  \cellcolor[HTML]{c0c0c0} &
  \multicolumn{4}{c|}{\cellcolor[HTML]{c0c0c0}\textit{\textbf{Weibo 16}}} &
  \multicolumn{4}{c|}{\cellcolor[HTML]{c0c0c0}\textit{\textbf{Weibo 20}}} &
  \multicolumn{4}{c|}{\cellcolor[HTML]{c0c0c0}\textbf{Sun-MM}} \\ \cline{3-14} 
\rowcolor[HTML]{c0c0c0} 
\multicolumn{1}{|l|}{\multirow{-2}{*}{\cellcolor[HTML]{c0c0c0}\textbf{Models}}} &
  \multirow{-2}{*}{\cellcolor[HTML]{c0c0c0}\textbf{Splits}} &
  \multicolumn{1}{l|}{\cellcolor[HTML]{c0c0c0}\textit{\textbf{Acc.}}} &
  \multicolumn{1}{c|}{\cellcolor[HTML]{c0c0c0}\textit{\textbf{P}}} &
  \multicolumn{1}{c|}{\cellcolor[HTML]{c0c0c0}\textit{\textbf{R}}} &
  \textit{\textbf{F1}} &
  \multicolumn{1}{l|}{\cellcolor[HTML]{c0c0c0}\textit{\textbf{Acc.}}} &
  \multicolumn{1}{c|}{\cellcolor[HTML]{c0c0c0}\textit{\textbf{P}}} &
  \multicolumn{1}{c|}{\cellcolor[HTML]{c0c0c0}\textit{\textbf{R}}} &
  \textit{\textbf{F1}} &
  \multicolumn{1}{l|}{\cellcolor[HTML]{c0c0c0}\textit{\textbf{Acc.}}} &
  \multicolumn{1}{c|}{\cellcolor[HTML]{c0c0c0}\textit{\textbf{P}}} &
  \multicolumn{1}{c|}{\cellcolor[HTML]{c0c0c0}\textit{\textbf{R}}} &
  \textit{\textbf{F1}} \\ \hline
 &
  \textit{Random} &
  \multicolumn{1}{c|}{\cellcolor[HTML]{EFEFEF}0.887} &
  \multicolumn{1}{c|}{\cellcolor[HTML]{EFEFEF}0.889} &
  \multicolumn{1}{c|}{\cellcolor[HTML]{EFEFEF}0.887} &
  \cellcolor[HTML]{EFEFEF}0.887 &
  \multicolumn{1}{c|}{\cellcolor[HTML]{EFEFEF}0.773} &
  \multicolumn{1}{c|}{\cellcolor[HTML]{EFEFEF}0.801} &
  \multicolumn{1}{c|}{\cellcolor[HTML]{EFEFEF}0.773} &
  \cellcolor[HTML]{EFEFEF}0.768 &
  \multicolumn{1}{c|}{\cellcolor[HTML]{EFEFEF}0.707} &
  \multicolumn{1}{c|}{\cellcolor[HTML]{EFEFEF}\bf 0.663} &
  \multicolumn{1}{c|}{\cellcolor[HTML]{EFEFEF}0.510} &
  \cellcolor[HTML]{EFEFEF}0.439 \\ \cline{2-14} 
 &
  \textit{Forward} &
  \multicolumn{1}{c|}{\textbf{0.936}} &
  \multicolumn{1}{c|}{\textbf{0.944}} &
  \multicolumn{1}{c|}{\textbf{0.936}} &
  \textbf{0.936} &
  \multicolumn{1}{c|}{0.699} &
  \multicolumn{1}{c|}{0.753} &
  \multicolumn{1}{c|}{0.699} &
  0.681 &
  \multicolumn{1}{c|}{0.701} &
  \multicolumn{1}{c|}{0.602} &
  \multicolumn{1}{c|}{0.507} &
  0.434 \\ \cline{2-14} 
\multirow{-3}{*}{\textbf{\begin{tabular}[c]{@{}c@{}}SVM-TS\\ w/o SP\end{tabular}}} &
  \textit{Reverse} &
  \multicolumn{1}{c|}{0.683} &
  \multicolumn{1}{c|}{0.698} &
  \multicolumn{1}{c|}{0.684} &
  0.678 &
  \multicolumn{1}{c|}{\textbf{0.831}} &
  \multicolumn{1}{c|}{\textbf{0.837}} &
  \multicolumn{1}{c|}{\textbf{0.831}} &
  \textbf{0.830} &
  \multicolumn{1}{c|}{\textbf{0.713}} &
  \multicolumn{1}{c|}{0.655} &
  \multicolumn{1}{c|}{\textbf{0.52}} &
  \textbf{0.453} \\ \hline
 &
  \textit{Random} &
  \multicolumn{1}{c|}{\cellcolor[HTML]{EFEFEF}0.929} &
  \multicolumn{1}{c|}{\cellcolor[HTML]{EFEFEF}0.930} &
  \multicolumn{1}{c|}{\cellcolor[HTML]{EFEFEF}0.929} &
  \cellcolor[HTML]{EFEFEF}0.929 &
  \multicolumn{1}{c|}{\cellcolor[HTML]{EFEFEF}0.925} &
  \multicolumn{1}{c|}{\cellcolor[HTML]{EFEFEF}0.926} &
  \multicolumn{1}{c|}{\cellcolor[HTML]{EFEFEF}0.925} &
  \cellcolor[HTML]{EFEFEF}0.925 &
  \multicolumn{1}{c|}{\cellcolor[HTML]{EFEFEF} \textbf{0.726}} &
  \multicolumn{1}{c|}{\cellcolor[HTML]{EFEFEF}\bf 0.674} &
  \multicolumn{1}{c|}{\cellcolor[HTML]{EFEFEF}\bf 0.691}  &
  \cellcolor[HTML]{EFEFEF}\bf 0.681 \\ \cline{2-14} 
 &
  \textit{Forward} &
  \multicolumn{1}{c|}{\textbf{0.938}} &
  \multicolumn{1}{c|}{\textbf{0.935}} &
  \multicolumn{1}{c|}{\textbf{0.934}} &
  \textbf{0.935} &
  \multicolumn{1}{c|}{0.851} &
  \multicolumn{1}{c|}{0.856} &
  \multicolumn{1}{c|}{0.851} &
  0.851 &
  \multicolumn{1}{c|}{0.655} &
  \multicolumn{1}{c|}{0.521} &
  \multicolumn{1}{c|}{0.514} &
  0.505 \\ \cline{2-14} 
\multirow{-3}{*}{\textbf{\begin{tabular}[c]{@{}c@{}}H-Trans / Hybrid \\ w/o SP\end{tabular}}} &
  \textit{Reverse} &
  \multicolumn{1}{c|}{0.730} &
  \multicolumn{1}{c|}{0.795} &
  \multicolumn{1}{c|}{0.732} &
  0.715 &
  \multicolumn{1}{c|}{\textbf{0.944}} &
  \multicolumn{1}{c|}{\textbf{0.945}} &
  \multicolumn{1}{c|}{\textbf{0.944}} &
  \textbf{0.944} &
  \multicolumn{1}{c|}{0.623} &
  \multicolumn{1}{c|}{0.516} &
  \multicolumn{1}{c|}{0.514} &
  0.514 \\ \hline
\end{tabular}
}
\caption{Ablation study of Weibo 16 \& 20 and Sun-MM datasets across three different data split strategies. Cells in \textbf{bold} indicate the best results from all models.}
\label{tab:results_4}
\end{table*}

\begin{table}[!t]
\small
\centering
\resizebox{\columnwidth}{!}{%
\begin{tabular}{|c|l|c|c|c|}
\hline
\rowcolor[HTML]{c0c0c0} 
\textbf{Dataset} &
  \multicolumn{1}{c|}{\cellcolor[HTML]{c0c0c0}\textbf{Splits}} &
  \textbf{IOU} &
  \textbf{DICE} &
  \textbf{Acc.} \\ \hline
 &
  \textit{Random} &
  \cellcolor[HTML]{EFEFEF}19.6 &
  \cellcolor[HTML]{EFEFEF}23.2 &
  \cellcolor[HTML]{EFEFEF}0.615 \\ \cline{2-5} 
 &
  \textit{Forward} &
  11.2 &
  13.8 &
  0.366 \\ \cline{2-5} 
\multirow{-3}{*}{\textbf{Twitter 15}} &
  \textit{Backward} &
  11.7 &
  14.5 &
  0.367 \\ \hline
 &
  \textit{Random} &
  \cellcolor[HTML]{EFEFEF}17.1 &
  \cellcolor[HTML]{EFEFEF}20.3 &
  \cellcolor[HTML]{EFEFEF}0.598 \\ \cline{2-5} 
 &
  \textit{Forward} &
  9.9 &
  12.3 &
  0.380 \\ \cline{2-5} 
\multirow{-3}{*}{\textbf{Twitter 16}} &
  \textit{Backward} &
  10.6 &
  13.1 &
  0.428 \\ \hline
 &
  \textit{Random} &
  \cellcolor[HTML]{EFEFEF}28.4 &
  \cellcolor[HTML]{EFEFEF}32.8 &
  \cellcolor[HTML]{EFEFEF}0.918 \\ \cline{2-5} 
 &
  \textit{Forward} &
  23.5 &
  28.4 &
  0.892 \\ \cline{2-5} 
\multirow{-3}{*}{\textbf{\begin{tabular}[c]{@{}c@{}}Weibo 16\\ Source\\ Post\end{tabular}}} &
  \textit{Backward} &
  22.3 &
  27.2 &
  0.812 \\ \hline
 &
  \textit{Random} &
  \cellcolor[HTML]{EFEFEF}26.2 &
  \cellcolor[HTML]{EFEFEF}30.2 &
  \cellcolor[HTML]{EFEFEF}0.920 \\ \cline{2-5} 
 &
  \textit{Forward} &
  20.9 &
  24.6 &
  0.738 \\ \cline{2-5} 
\multirow{-3}{*}{\textbf{\begin{tabular}[c]{@{}c@{}}Weibo 20\\ Source\\ Post\end{tabular}}} &
  \textit{Backward} &
  21.8 &
  26.2 &
  0.898 \\ \hline
 &
  \textit{Random} &
  \cellcolor[HTML]{EFEFEF}23.5 &
  \cellcolor[HTML]{EFEFEF}27.2 &
  \cellcolor[HTML]{EFEFEF}0.839 \\ \cline{2-5} 
 &
  \textit{Forward} &
  14.0 &
  16.6 &
  0.708 \\ \cline{2-5} 
\multirow{-3}{*}{\textbf{Sun-MM}} &
  \textit{Backward} &
  13.4 &
  17.3 &
  0.807 \\ \hline
 &
  \textit{Random} &
  \multicolumn{1}{l|}{\cellcolor[HTML]{EFEFEF}26.7} &
  \multicolumn{1}{l|}{\cellcolor[HTML]{EFEFEF}31.3} &
  \multicolumn{1}{l|}{\cellcolor[HTML]{EFEFEF}0.929} \\ \cline{2-5} 
 &
  \textit{Forward} &
  \multicolumn{1}{l|}{26.2} &
  \multicolumn{1}{l|}{31.2} &
  \multicolumn{1}{l|}{0.938} \\ \cline{2-5} 
\multirow{-3}{*}{\textbf{\begin{tabular}[c]{@{}c@{}}Weibo 16\\ Comment\end{tabular}}} &
  \textit{Backward} &
  \multicolumn{1}{l|}{25.4} &
  \multicolumn{1}{l|}{30.0} &
  \multicolumn{1}{l|}{0.730} \\ \hline
 &
  \textit{Random} &
  \multicolumn{1}{l|}{\cellcolor[HTML]{EFEFEF}26.0} &
  \multicolumn{1}{l|}{\cellcolor[HTML]{EFEFEF}30.3} &
  \multicolumn{1}{l|}{\cellcolor[HTML]{EFEFEF}0.925} \\ \cline{2-5} 
 &
  \textit{Forward} &
  \multicolumn{1}{l|}{25.5} &
  \multicolumn{1}{l|}{30.1} &
  \multicolumn{1}{l|}{0.851} \\ \cline{2-5} 
\multirow{-3}{*}{\textbf{\begin{tabular}[c]{@{}c@{}}Weibo 20\\ Comment\end{tabular}}} &
  \textit{Backward} &
  \multicolumn{1}{l|}{24.8} &
  \multicolumn{1}{l|}{28.5} &
  \multicolumn{1}{l|}{0.944} \\ \hline
\end{tabular}%
}
\caption{Textual similarity betweet training and test sets using random and temporal data splits.}
\label{tab:similarity}
\end{table}

\subsection{Similarity Analysis}
\label{section:Similarity}
This section explores the impact of data split strategies on the content and contextual information in the respective training and test sets.
We investigate whether a decrease in model predictive performance occurs due to variations between the two subsets used for training and testing, and whether the difference in performance lessens as the datasets become more similar to each other.
\textbf{Source Post} Similar to \citet{kochkina2023evaluating,mu2023examining}, we first measure the difference in textual similarity between training and test sets generated using random and chronological data splits using two standard matrices with ranges from 0 to 1.
\paragraph{Intersection over Union (IoU) \citep{tanimoto1958elementary}}
\begin{equation}
    IoU = \frac{|V^{Train} \cap V^{Test}|}{|V^{Train} \cup  V^{Test}|}
\end{equation}

\paragraph{DICE coefficient (DICE) \citep{dice1945measures}}
\begin{equation}
    DICE = \frac{2 \times |V^{Train} \cap V^{Test}|}{|V^{Train}| + |V^{Test}|}
\end{equation}
where $V^{Train}$ and $V^{Test}$ refer to the set of unique words from training and test sets; and $|V^{Train} \cap V^{Test}|$ and $|V^{Train} \cup  V^{Test}|$ indicate the number of unique words that appear in the \textit{intersection} and \textit{union} of training and test sets respectively. When the two sets have no shared vocabulary list, the IoU and DICE values will be 0, while if they are identical, the IoU and DICE values will be 1.

We display the similarity of the source posts between training and test sets using different data split strategies in Table \ref{tab:similarity}. Additionally, we provide the accuracy of the BERT model (which takes only the source post as input) for each dataset as a reference.

We demonstrate that using random splits leads to significantly higher IoU and DICE values ($t$-test, $p$ < 0.001), indicating greater similarities between the training and test sets compared to both forward and backward chronological splits. This suggests that rumors with similar content, resulting from temporal concept drift, appear in both training and test sets when employing random data splits. Additionally, we discover a positive correlation (using Pearson's Test) between model accuracy and the similarity distance of training and test sets, as measured by both IOU (Pearsons' $r$ = 0.865, $p$ < 0.05) and DICE (Pearsons' $r$ = 0.879, $p$ < 0.001) values. In other words, higher textual similarities correspond to better classifier performance.

\begin{figure*}[!t]
\centering
  \includegraphics[scale=0.35]{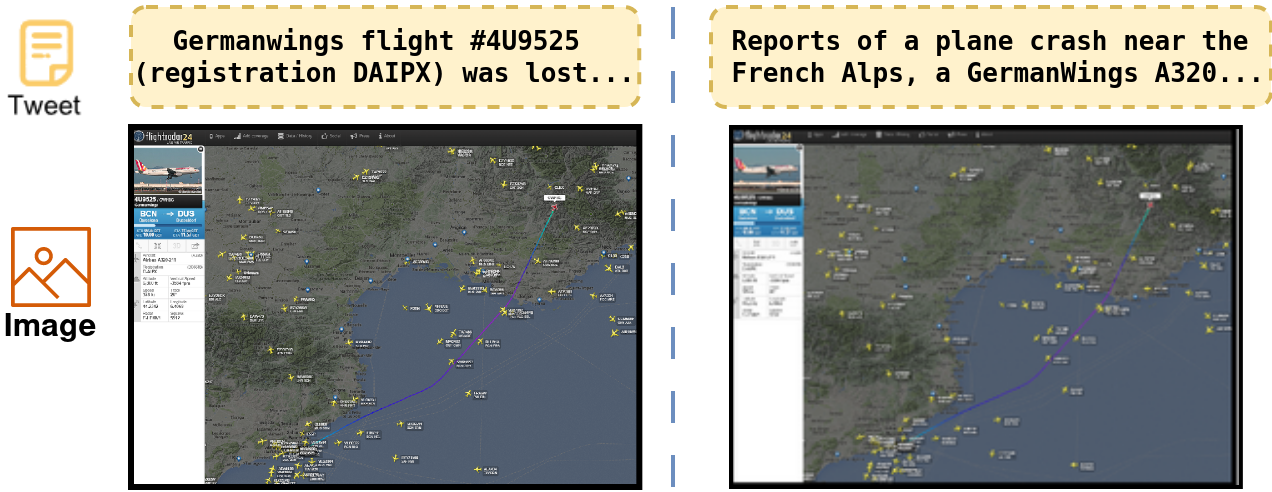}
  \caption{Two rumors from the Sun-MM Dataset related to the 'plane crash' event contain similar images and were published during a comparable time period. More examples are displayed in the Appendix (see Figure \ref{fig:sun-mm-more-exampels}).}
  \label{fig:sun-mm}
\end{figure*}

\paragraph{User Profile Attributes}
We use cosine similarity to assess the difference between the mean values of user profile attributes from the training and test sets. However, we do not observe a significant difference in cosine similarity values when using both random and chronological data splits, as all rumor speaders are unique across all datasets.
\paragraph{Comments}
Considering the comparable model performance, with accuracy of up to 93.8\% and 94.4\% on Weibo 16 and Weibo 20, when using only comments as input, we hypothesize that the comments from the two classes are significantly different. To identify the difference in comments that distinguish between rumors and non-rumors in Weibo 16 and Weibo 20, we employ the univariate Pearson's correlation test \citep{schwartz2013personality}.
We observe that there is a large amount of words related to debunking rumors (e.g., `false', `really?', and `truth') in the comments associated with false rumors on both Weibo 16 and Weibo 20. On the other hand, comments associated with non-rumors are more words related to the daily life of the public. Note that non-rumors in Weibo datasets are collected from mainstream media accounts. 
\paragraph{Images}
Ablation study results (see Table \ref{tab:results_4}) show that only the ViT model, which uses images alone as input, is affected by the temporal data splits (i.e., the deterioration of model performance). We further explore the Sun-MM dataset and uncover that rumors with similar content are usually posted with similar images. 
We show examples in Figure \ref{fig:sun-mm} . Note that similar semantic objects (e.g., entitles \citep{sun-etal-2021-inconsistency-matters}) can be extracted from similar images, which can impact the accuracy of the model.

\section{How do we properly use static datasets?}
\label{suggestions}
Apart from prioritizing skewed methodologies solely for achieving high accuracy on rumor detection datasets, it is essential to develop a deeper comprehension of the protocol we employ and generate significant insights. Given the limitations raised by our experiments, we make the following practical suggestions for developing new rumor detection systems on static datasets:

\begin{itemize}
    \item For practical applications that aim to detect \textbf{unseen rumors}, it is essential to consider chronological splits when evaluating all rumor detection approaches on static datasets, in addition to standard random splits. By using forward and backward chronological splits, we can assess the ability of the rumor classifiers to handle both earlier and older unseen rumors. 

    \item Considering that temporalities (i.e., the temporal concentration of rumor topics) typically occur in widely used rumor detection datasets (e.g., Twitter 15\&16 and Weibo 16 \citep{ma2016detecting,ma2017detect}), one can apply an additional data pre-processing measure to filter out rumor events with multiple posts. For instance, using out-of-the-box methods such as Levenshtein distance \citep{levenshtein1966binary} and BERTopic \citep{grootendorst2022bertopic}, we identified a total of 9 similar rumors that resemble the false rumor depicted in Figure \ref{fig:ex}. After conducting a more in-depth error analysis on the predictions generated by the H-Trans model, which has demonstrated the highest predictive performance on Weibo 16, we discovered that the models can accurately classify all of these rumors in the test set when employing random data splits.

    \item Current evaluation metrics, such as accuracy and F1-measure, are unable to accurately assess the true capability of rumor classifiers in detecting unseen rumors. Therefore, there is a need for new measures to evaluate the accuracy of model predictions for unknown rumors. For example, one can calculate the accuracy of a rumor detection system by excluding known rumors (i.e., similar rumors appearing in the training set) from the test set.

    \item Given the limitations of the current pipeline that relies solely on static datasets, we argue that evaluation models should not be restricted to such datasets. By leveraging the consistent format of datasets collected from the same platform (as shown in Table \ref{tab:datasets}), for example, one can explore \textbf{broader temporalities} by training a rumor classifier on Twitter 15 and evaluating its performance on Twitter 16. This protocol enables a more comprehensive examination of the generalizability of rumor detection systems, which is crucial for their practical applications in the real world \citep{moore2018bringing,yin2021towards,kochkina2023evaluating}.

\end{itemize}

\section{Conclusion}
In this paper, we evaluate the limitations of existing widely used rumor detection models trained on static datasets. Through empirical analysis, we demonstrate that the use of chronological splits significantly diminishes the predictive performance of widely-used rumor detection models. To better understand the causes behind these limitations, we conduct a fine-grained similarity analysis and an ablations study. Finally, we provide practical recommendations for future research in the advancement of new rumor detection systems.
\paragraph{Limitations and Future Work}
We conducted an empirical study on current rumor detection models, utilizing both the source post and \textbf{standard contextual information} (such as comments, images, and user profile attributes) as input. However, previous research has employed hidden features, such as sentiment and entities, which can be extracted from the source post and contextual information \citep{rao-etal-2021-stanker, sun-etal-2021-inconsistency-matters}. We consider this as future work and aim to explore additional feature settings.
Besides, the current work is limited to English and Chinese, and we acknowledge that further research into more multilingual datasets should be considered in the future.
\section*{Ethics Statement}
Our work has been approved by the Research Ethics Committee of the University of Sheffield, and complies with the data policies of Twitter\footnote{\url{https://developer.twitter.com/en/docs/twitter-api}} and Weibo\footnote{\url{https://open.weibo.com}}. All datasets are obtained through the links provided in the source papers.

\section*{Acknowledgments}
This research is supported by an EU Horizon 2020 grant (agreement no.871042) (``So-BigData++: European Integrated Infrastructure for Social Mining and BigData Analytics'').\footnote{\url{http://www.sobigdata.eu}}

\section*{References}
\bibliographystyle{lrec-coling2024-natbib}
\bibliography{lrec-coling2024-example}

\label{sec:appendix}
\section*{Appendix}

\begin{figure*}[!t]
\centering
  \includegraphics[scale=0.32]{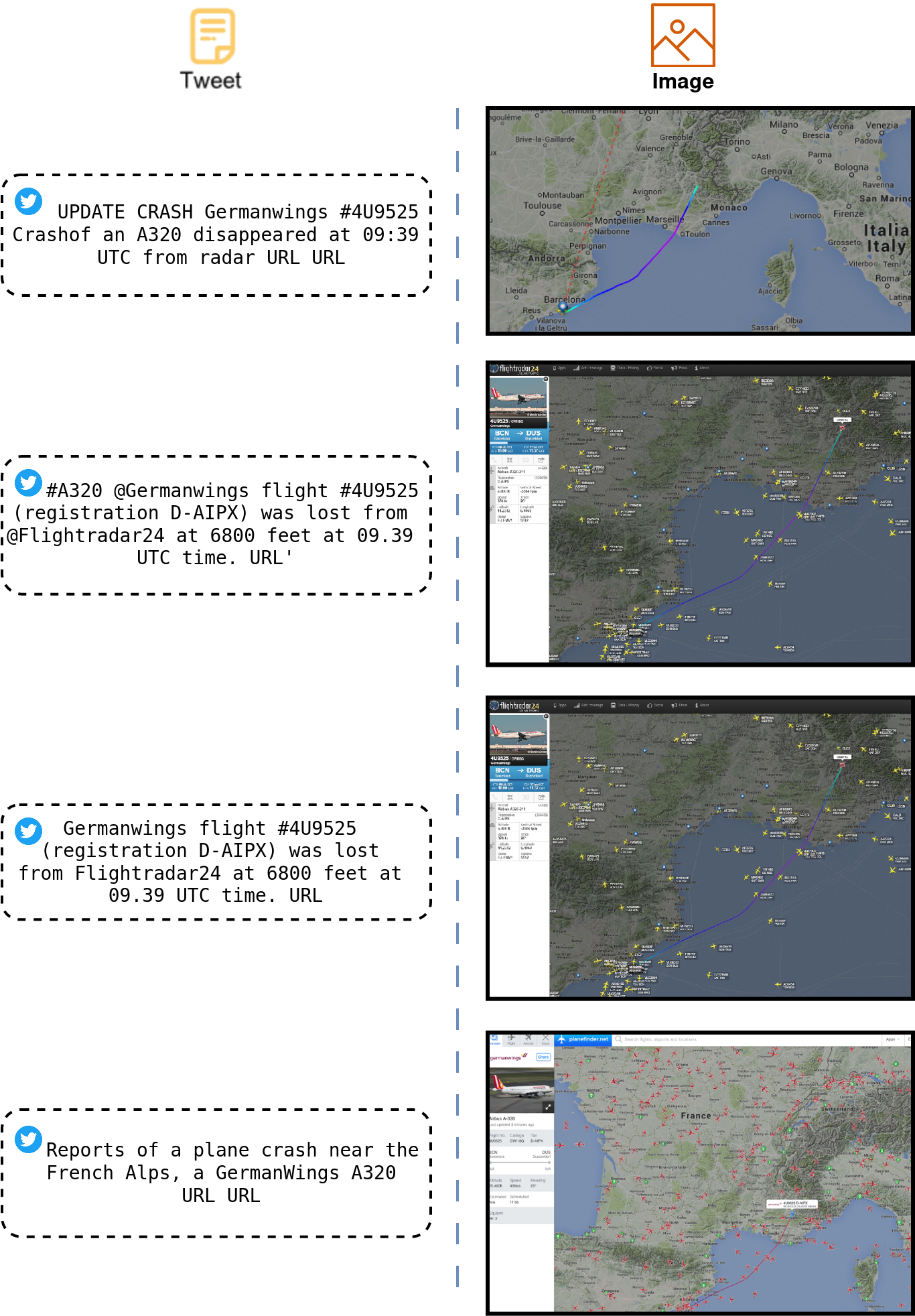}
  \caption{Four pairs of rumors related to the 'plane crash' event (from the Sun-MM Dataset) contain similar images and were published during a comparable time period.}
  \label{fig:sun-mm-more-exampels}
\end{figure*}
\end{document}